\documentstyle[11pt,lingmacros]{article}

\input{psfig}

\makeatletter

\def\@citex[#1]#2{\if@filesw\immediate\write\@auxout{\string\citation{#2}}\fi
  \def\@citea{}\@cite{\@for\@citeb:=#2\do
    {\@citea\def\@citea{;\penalty\@m\ }\@ifundefined
       {b@\@citeb}{{\bf ?}\@warning
       {Citation `\@citeb' on page \thepage \space undefined}}%
{\csname b@\@citeb\endcsname}}}{#1}}
\let\@internalcite\cite
\def\cite{\def\citename##1{##1, }\@internalcite}
\def\shortcite{\def\citename##1{}\@internalcite}
\def\newcite{\leavevmode\def\citename##1{{##1} (}\@internalciteb}

\def\@citexb[#1]#2{\if@filesw\immediate\write\@auxout{\string\citation{#2}}\fi
  \def\@citea{}\@newcite{\@for\@citeb:=#2\do
    {\@citea\def\@citea{;\penalty\@m\ }\@ifundefined
       {b@\@citeb}{{\bf ?}\@warning
       {Citation `\@citeb' on page \thepage \space undefined}}%
\hbox{\csname b@\@citeb\endcsname}}}{#1}}
\def\@internalciteb{\@ifnextchar
[{\@tempswatrue\@citexb}{\@tempswafalse\@citexb[]}}

\def\@newcite#1#2{{#1\if@tempswa, #2\fi)}}

\def\@biblabel#1{\def\citename##1{##1}[#1]\hfill}

\def\@cite#1#2{({#1\if@tempswa , #2\fi})}

\def\thebibliography#1{\vskip\parskip%
\vskip\baselineskip%
\def\baselinestretch{1}%
\ifx\@currsize\normalsize\@normalsize\else\@currsize\fi%
\vskip-\parskip%
\vskip-\baselineskip%
\section*{References\@mkboth
 {References}{References}}\list
 {}{\setlength{\labelwidth}{0pt}\setlength{\leftmargin}{\parindent}
 \setlength{\itemindent}{-\parindent}}
 \def\newblock{\hskip .11em plus .33em minus -.07em}
 \sloppy\clubpenalty4000\widowpenalty4000
 \sfcode`\.=1000\relax}

\def\thesourcebibliography#1{\vskip\parskip%
\vskip\baselineskip%
\def\baselinestretch{1}%
\ifx\@currsize\normalsize\@normalsize\else\@currsize\fi%
\vskip-\parskip%
\vskip-\baselineskip%
\section*{Sources of Attested Examples\@mkboth
 {Sources of Attested Examples}{Sources of Attested Examples}}\list
 {}{\setlength{\labelwidth}{0pt}\setlength{\leftmargin}{\parindent}
 \setlength{\itemindent}{-\parindent}}
 \def\newblock{\hskip .11em plus .33em minus -.07em}
 \sloppy\clubpenalty4000\widowpenalty4000
 \sfcode`\.=1000\relax}

\def\@lbibitem[#1]#2{\item[]\if@filesw
      { \def\protect##1{\string ##1\space}\immediate
        \write\@auxout{\string\bibcite{#2}{#1}}\fi\ignorespaces}}

\def\@bibitem#1{\item\if@filesw \immediate\write\@auxout
       {\string\bibcite{#1}{\the\c@enumi}}\fi\ignorespaces}

\makeatother

\author{Andrew Kehler \\ Harvard University \\
Aiken Computation Laboratory
\\ 33 Oxford Street \\ Cambridge, MA 02138 \\ kehler@das.harvard.edu}
\title{\vspace{-.3in} \LARGE\bf COMMON TOPICS AND COHERENT SITUATIONS:
 INTERPRETING ELLIPSIS IN THE CONTEXT OF DISCOURSE INFERENCE
\\ \vspace{.12in} \small {\it (To appear in the Proceedings of ACL-94)}
}

\begin{document}

\maketitle

\begin{abstract}
It is claimed that a variety of facts concerning ellipsis, event
reference, and interclausal coherence can be explained by two features
of the linguistic form in question: (1) whether the form  leaves
behind an empty constituent in the syntax, and (2) whether the
form is anaphoric in the semantics.  It is proposed that these
features interact with one of two types of discourse inference, namely
{\it Common Topic} inference and {\it Coherent Situation} inference.
The differing ways in which these types of inference utilize syntactic
and semantic representations predicts phenomena for which it
is otherwise difficult to account.
\end{abstract}

\bibliographystyle{acl}

\section{Introduction}
Ellipsis is pervasive in natural language, and hence has received much
attention within both computational and theoretical linguistics.
However, the conditions under which a representation of an utterance
may serve as a suitable basis for interpreting subsequent elliptical
forms remain poorly understood; specifically, past attempts to
characterize these processes within a single traditional module of
language processing (e.g., considering either syntax, semantics, or
discourse in isolation) have failed to account for all of
the data.  In this paper, we claim that a variety of facts concerning
ellipsis resolution, event reference, and interclausal coherence can be
explained by the interaction between the syntactic and semantic properties
of the form in question and the type of discourse inference operative
in establishing the coherence of the antecedent and elided clauses.

In the next section, we introduce the facts concerning gapping,
VP-ellipsis, and non-elliptical event reference that we seek to explain.
In Section~3, we categorize elliptical
and event referential forms according to two features: (1) whether the
expression leaves behind an empty constituent in the syntax, and (2)
whether the expression is anaphoric in the semantics.  In
Section~4 we describe two types of
discourse inference, namely {\it Common Topic} inference and {\it
Coherent Situation} inference, and make a specific proposal concerning
the interface between these and the syntactic and semantic
representations they utilize.  In Section~5, we
show how this proposal accounts for the data presented in
Section~2.  We contrast the account with
relevant past work in Section~6, and conclude in
Section~7.

\section{Ellipsis and Interclausal Coherence}
\label{motivation-section}

It has been noted in previous work that the felicity of certain forms of
ellipsis is dependent on the type of coherence relationship extant
between the antecedent and elided clauses \cite{LevinPri:82,Kehler:93b}.
In this section we review the relevant facts for two such forms of
ellipsis, namely {\it gapping} and {\it VP-ellipsis}, and also compare
these with facts concerning non-elliptical event reference.

Gapping is characterized by an antecedent sentence (henceforth called the
{\it source} sentence) and the elision of all but two constituents
(and in limited circumstances, more than two constituents) in one
or more subsequent {\it target} sentences, as exemplified in sentence
(\ref{gap1}):

\enumsentence{Bill became upset, and Hillary
angry. \label{gap1}}
We are concerned here with a
particular fact about gapping noticed by Levin and Prince
\shortcite{LevinPri:82}, namely that gapping is
acceptable only with the purely conjunctive {\it symmetric} meaning of
{\it and} conjoining the clauses, and not with its causal {\it
asymmetric} meaning (paraphraseable by ``and as a result'').  That is,
while either of sentences (\ref{gap1}) or (\ref{ungap1}) can have the
purely conjunctive reading, only sentence (\ref{ungap1}) can be
understood to mean that Hillary's becoming angry was caused by or came
as a result of Bill's becoming upset.

\enumsentence{Bill became upset, and Hillary became
angry. \label{ungap1}}
This can be seen by embedding each of these examples in a context that
reinforces one of the meanings.  For instance, gapping is felicitous
in passage (\ref{sym-gap}), where context supports the symmetric
reading, but is infelicitous in passage (\ref{assym-gap}) under the
intended causal meaning of {\it and}.\footnote{This behavior is not
limited to the conjunction {\it and}; a similar distinction holds
between symmetric and asymmetric uses of {\it or} and {\it but}.  See
Kehler \shortcite{Kehler:94a} for further discussion. }

\enumsentence{The Clintons want to get the national debate
focussed on health care, and are getting annoyed because the media is
preoccupied with Whitewater.  When a reporter recently asked a
Whitewater question at a health care rally, Bill became
upset, and Hillary became/$\emptyset$ angry.
\label{sym-gap} }

\enumsentence{Hillary has been getting annoyed at Bill for his inability
to deflect controversy and do damage control.  She has repeatedly told
him that the way to deal with Whitewater is to play it down and not to
overreact.  When a reporter recently asked a Whitewater question at a
health care rally, Bill became upset, and (as a result) Hillary
became/\# $\emptyset$ angry.
\label{assym-gap} }
The common stipulation within the literature stating that gapping applies to
coordinate structures and not to subordinate ones does not account
for why any coordinated cases are unacceptable.

VP-ellipsis is characterized by an initial
{\it source} sentence, and a subsequent {\it target} sentence with a
bare auxiliary indicating the elision of a verb phrase:

\enumsentence{Bill became upset, and Hillary did too. \label{foo3}}

The distribution of VP-ellipsis has also been shown to be sensitive to
the coherence relationship extant between the source and target
clauses, but in a different respect.  In a previous paper
\cite{Kehler:93b}, five contexts for VP-ellipsis
were examined to determine whether the representations
retrieved are syntactic or semantic in nature.  Evidence was given that
VP-ellipsis copies syntactic representations in what was termed {\it
parallel} constructions (predicting the unacceptability
of the voice mismatch in example (\ref{ell1}) and nominalized source in
example (\ref{ell3})), but copies semantic representations in {\it
non-parallel} constructions (predicting the acceptability of the voice
mismatch in example (\ref{ell2}) and the nominalized source in example
(\ref{ell4})):\footnote{These examples have been taken or adapted from
Kehler \shortcite{Kehler:93b}.  The phrases shown in brackets indicate the
elided material under the intended interpretation.}

\enumsentence{\# The decision was reversed by the FBI, and the ICC did
too.  [ reverse the decision ]  \label{ell1}}

\enumsentence{In March, four fireworks manufacturers asked that the decision be
reversed, and on Monday the ICC did. [ reverse the decision ]
 \label{ell2}}

\enumsentence{\# This letter provoked a response from Bush, and
Clinton did too.  [ respond ] \label{ell3} }

\enumsentence{This letter was meant to provoke a response from
Clinton, and so he did. [ respond ] \label{ell4}}
These examples are analogous with the gapping cases in that
constraints against mismatches of syntactic form hold for the
symmetric (i.e., parallel) use of {\it and}  in examples
(\ref{ell1}) and (\ref{ell3}), but not the asymmetric (i.e.,
non-parallel) meaning in examples (\ref{ell2}) and
(\ref{ell4}).  In fact, it appears that gapping is felicitous in those
constructions where VP-ellipsis requires a syntactic antecedent,
whereas gapping is infelicitous in cases where VP-ellipsis requires
only a suitable semantic antecedent.  Past approaches to VP-ellipsis
that operate within a single module of language processing fail to
make the distinctions necessary to account for these differences.

Sag and Hankamer \shortcite{Sag:84a} note that while elliptical
sentences such as (\ref{ell1}) are unacceptable because of a voice
mismatch, similar examples with non-elided event referential forms
such as {\it do it} are much more acceptable:

\enumsentence{  \label{non-ell1}
The decision was reversed by the FBI, and the ICC did
it too.  [ reverse the decision ] }
An adequate theory of ellipsis and event reference must
account for this distinction.

In sum, the felicity of both gapping and VP-ellipsis appears to be
dependent on the type of coherence relation extant between the source
and target clauses.  Pronominal event reference, on the other hand,
appears not to display this dependence.  We seek to account for these
facts in the sections that follow.

\section{Syntax and Semantics of Ellipsis and Event Reference}
\label{syn-sem-section}

In this section we characterize the forms being addressed in terms of
two features: (1) whether the form leaves behind an empty constituent
in the syntax, and (2) whether the form is anaphoric in the semantics.
In subsequent sections, we show how the distinct mechanisms for
recovering these types of missing information interact with two types
of discourse inference to predict the phenomena noted in the previous
section.

\begin{figure}[ht]
\centerline{\psfig{figure=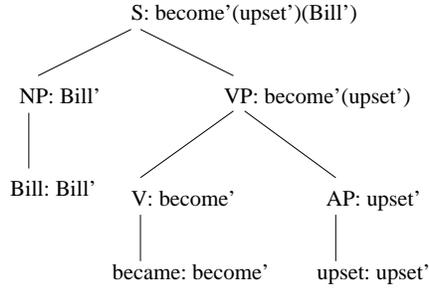,height=1.5in,width=2.25in}}

\caption{Syntactic and Semantic Representations for {\it Bill
became upset.}}
\label{rep-figure}
\end{figure}

We illustrate the relevant syntactic and semantic properties of these
forms using the version of Categorial Semantics described in Pereira
\shortcite{Pereira:90}.  In the Montagovian tradition, semantic
representations are compositionally generated in correspondence with
the constituent modification relationships manifest in the syntax;
predicates are curried.  Traces are associated with assumptions which
are subsequently discharged by a suitable construction.
Figure~\ref{rep-figure} shows the representations for the sentence
{\it Bill became upset}; this will serve as the initial source clause
representation for the examples that follow.\footnote{We will ignore
the tense of the predicates for ease of exposition.}

\begin{figure}[ht]

\centerline{\psfig{figure=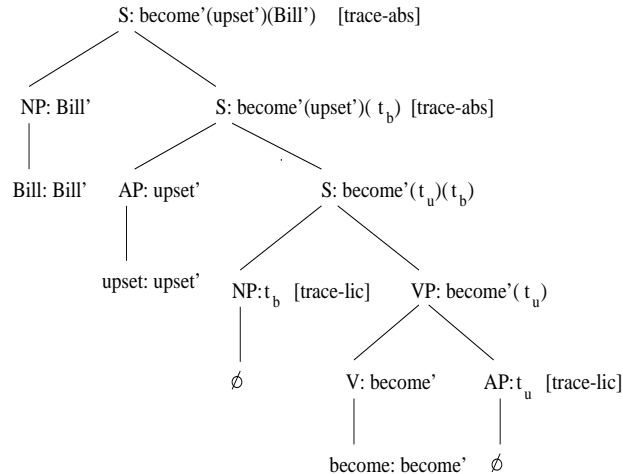,height=2.5in,width=3.25in}}

\caption{Syntactic and Semantic Representations for {\it Bill
became upset} after fronting.}
\label{extrapose-figure}
\end{figure}

For our analysis of gapping, we follow Sag
\shortcite{Sag:76a} in hypothesizing that a post-surface-structure
level of syntactic representation is used as the basis for
interpretation.  In source clauses of gapping constructions,
constituents in the source that are parallel to the overt constituents
in the target are abstracted out of the clause
representation.\footnote{It has been noted that in gapping
constructions, contrastive accent is generally placed on parallel
elements in both the target and the source clauses, and that
abstracting these elements results in an ``open proposition'' that
both clauses share
\cite{Sag:76a,Prince:86,Steedman:90}.  This proposition needs to be
presupposed (or accommodated) for the gapping to be felicitous, for
instance, it would be infelicitous to open a conversation with
sentence such as (\ref{gap1}), whereas it is perfectly felicitous in
response to the question {\it How did the Clintons react?}.  Gapping
resolution can be characterized as the restoration of this open proposition in
the gapped clause.} For simplicity, we will assume that this
abstraction is achieved by fronting the constituents in the
post-surface-structure, although nothing much hinges on this; our
analysis is compatible with several possible mechanisms.  The
syntactic and semantic representations for the source clause of
example (\ref{gap1}) after fronting are shown in Figure
\ref{extrapose-figure}; the fronting leaves trace assumptions behind
that are discharged when combined with their antecedents.

\begin{figure}[ht]
\centerline{\psfig{figure=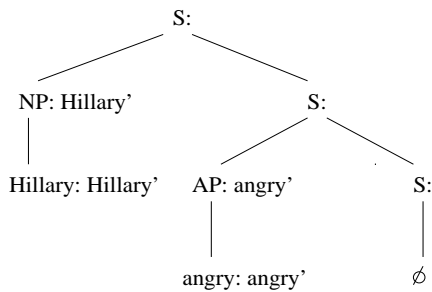,height=1.5in,width=2.25in}}
\caption{Syntactic and Semantic Representations for {\it Hillary angry}.}
\label{target-gap-figure}
\end{figure}

Target clauses in gapping constructions are therefore represented with
the overt constituents fronted out of an elided sentence node; for
instance the representation of the target clause in example
(\ref{gap1}) is shown in Figure~\ref{target-gap-figure} (the empty
node is indicated by $\phi$).
The empty constituent is reconstructed
by copying the embedded sentence from the source to the
target clause, along with parallel trace assumptions which are to be
bound within the target.  The semantics for this embedded sentence is
the open proposition that the two clauses share.  This semantics, we
claim, can only be recovered by copying the syntax, as gapping
does not result in an independently anaphoric expression in the
semantics.\footnote{This claim is supported by well-established facts
suggesting that gapping does not pattern with standard forms of
anaphora.  For instance, unlike VP-ellipsis and overt pronouns,
gapping cannot be cataphoric, and can only obtain its
antecedent from the immediately preceding clause.}  In fact, as can be
seen from Figure~\ref{target-gap-figure},  before
copying takes place there is no sentence-level semantics for gapped
clauses at all.

\begin{figure}[ht]

\centerline{\psfig{figure=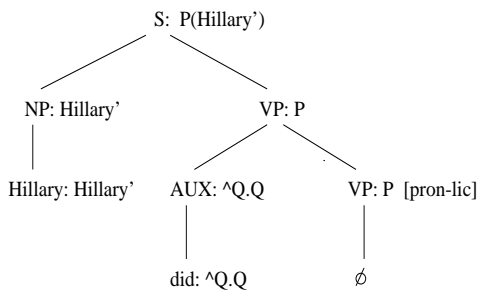,height=1.5in,width=2.5in}}

\caption{Syntactic and Semantic Representations for {\it Hillary did.}}
\label{ellip-fig}
\end{figure}

Like gapping, VP-ellipsis results in an empty constituent in the
syntax, in this case, a verb phrase.  However, unlike gapping,
VP-ellipsis also results in an independently anaphoric form in the
semantics.\footnote{Unlike gapping, VP-ellipsis patterns with
other types of anaphora, for instance it can be cataphoric and can
locate antecedents from clauses other than the most immediate one.}
Figure~\ref{ellip-fig} shows the representations for the clause
{\it Hillary did} (the anaphoric expression is indicated by $P$).

Given the representation in Figure~\ref{rep-figure} as the source, the
semantics for the missing VP may be
recovered in one of two ways.  The syntactic VP could be copied down
with its corresponding semantics, from which the semantics for the
complete sentence can be derived.  In this case, the anaphoric
expression is constrained to have the same semantics as the copied
constituent.  Alternatively, the anaphoric expression could be
resolved purely semantically, resulting in the discharge of the
anaphoric assumption $P$.  The higher-order unification method
developed by Dalrymple et al.\ \shortcite{Dalrympl:91a} could be used
for this purpose; in this case the sentence-level semantics is
recovered without copying any syntactic representations.

Event referential forms such as {\it do it}, {\it do that}, and {\it
do so} constitute full verb phrases in the syntax.  It has been often
noted \cite[inter alia]{HallidayHasan:76} that it is the main verb
{\it do} that is operative in these forms of anaphora, in contrast to
the auxiliary {\it do} operative in VP-ellipsis.\footnote{For
instance, other auxiliaries can appear in elided forms but cannot be
followed by {\it it}, {\it that}, or {\it so} as in example
(\ref{no1}), and a pronominal object to the main verb {\it do} cannot
refer to a state as VP-ellipsis can as in example (\ref{no2}).

\enumsentence{George was going to the golf course and Bill was $\emptyset$/(\#
it)/(\# that)/(\# so) too. \label{no1}}

\enumsentence{Bill dislikes George and Hillary does $\emptyset$/(\#
it)/(\# that)/(\# so) too. \label{no2}}}
It is the pronoun in event referential forms that is anaphoric; the fact that
the pronouns refer to events results from the type constraints imposed by the
main verb {\it do}.  Therefore, such forms are anaphoric in the
semantics, but do not leave behind an empty constituent in the syntax.

To summarize this section, we have characterized the forms being
addressed according to two features, a summary of which appears in
Table~\ref{summary-table}.
\begin{table}[ht]
\begin{center}
\begin{tabular}{||c|c|c||} \hline
Form  & Empty Node & Anaphoric \\
      & in Syntax         & in Semantics \\ \hline \hline
Gapping & $\surd$ &  \\ \hline
VP-Ellipsis & $\surd$ & $\surd$ \\ \hline
Event Reference &  & $\surd$ \\ \hline
\end{tabular}
\end{center}
\caption{Common Topic Relations}
\label{summary-table}
\end{table}
Whereas anaphoric forms in the semantics for these forms are
independently resolved, empty syntactic constituents in and of
themselves are not anaphoric, and thus may only be restored when some
independently-motivated process necessitates it.  In the section that
follows we outline two types of discourse inference, one of which
requires such restoration of empty constituents.

\section{Discourse Inference}
\label{disc-inference-section}

To be coherent, utterances within a discourse segment require more
than is embodied in their individual syntactic and semantic
representations alone; additional inter-utterance constraints must be
met.  Here we describe two types of inference used to enforce the
constraints that are imposed by coherence relations.
In each case, arguments to coherence relations take
the form of semantic representations retrieved by way of their
corresponding node(s) in the syntax; the operations performed on these
representations are dictated by the nature of the constraints imposed.  The two
types of inference are distinguished by the level in the syntax from
which these arguments are retrieved.\footnote{Hobbs
\shortcite{Hobbs:90}, following Hume \shortcite{Hume}, suggests a
classification of coherence relations into three broad categories,
namely {\it Resemblance}, {\it Cause or Effect}, and {\it Contiguity}
(Hume's terminology).  Here, {\it Resemblance} relations appear to
pattern well with those employing our Common Topic inference, and likewise
{\it Cause or effect} and {\it Contiguity} with our Coherent Situation
inference.}

\subsection{Common Topic Inference}

Understanding segments of utterances standing in a Common Topic
relation requires the determination of points of commonality
(parallelism) and departure (contrast) between sets of corresponding
entities and properties within the utterances.  This process is reliant
on performing comparison and generalization operations on the
corresponding representations \cite{Scha:88a,Hobbs:90,Prust:92,Asher:93}.
Table~\ref{common-topic-table} sketches definitions for some Common
Topic relations, some taken from and others adapted from Hobbs
\shortcite{Hobbs:90}.  In each case, the hearer is to understand the
relation by inferring $p_{0}(a_1,...,a_n)$ from sentence $S_0$ and
inferring $p_{1}(b_1,...,b_n)$ from sentence $S_1$ under the listed
constraints.\footnote{Following Hobbs, by  $a_i$ and $b_{i}$ being {\it
similar} we mean
that for some salient property $q_i$, $q_i(a_i)$ and $q_i(b_i)$ holds.
Likewise by {\it dissimilar} we mean that for some $q_i$,
$q_i(a_i)$ and $\neg q_i(b_i)$ holds.} In order to meet these constraints,
the identification of $p_{0}$ and $p_{1}$ may require arbitrary
levels of generalization from the relations explicitly stated in the
utterances.

\begin{table*}[ht]
\begin{center}
\begin{tabular}{||c|c|c||} \hline
Relation & Constraints & Conjunctions \\ \hline \hline
Parallel & $p_{0} = p_{1}$, $a_i$ and $b_i$ are similar & and   \\ \hline
Contrast  & (1) $p_{0} = \neg p_{1}$, $a_i$ and $b_i$ are similar & but   \\
 & (2) $p_{0} = p_{1}$, $a_i$ and $b_i$ are dissimilar for
some $i$ &    \\ \hline
Exemplification & $p_{0} = p_{1}$ ; $b_i \in a_i$ or $b_i \subset
a_i$  & for example   \\ \hline
Elaboration & $p_{0} = p_{1}$, $a_i = b_i$ & in other words   \\ \hline
\end{tabular}
\end{center}
\caption{Common Topic Relations}
\label{common-topic-table}
\end{table*}
 Examples of these relations are given in sentences (\ref{CT-examples}a-d).

\eenumsentence{  \label{CT-examples}
\item John organized rallies for Clinton, and Fred distributed pamphlets
for him.  (Parallel)
\item John supported Clinton, but Mary supported Bush.  (Contrast)
\item Young aspiring politicians usually support their party's
presidential candidate.  For instance,
John campaigned hard for Clinton in 1992. (Exemplification)
\item A young aspiring politician was arrested in Texas today.  John
Smith, 34, was nabbed in a Houston law firm while attempting to
embezzle funds for his campaign. (Elaboration)
}
Passage (\ref{CT-examples}a), for instance, is coherent under the
understanding that John and Fred have a common property, namely {\it
having done something to support Clinton}.  Passage (\ref{CT-examples}c) is
likewise coherent by virtue of the inferences resulting from
identifying parallel elements and properties, including that John is a
young aspiring politician and that he's a Democrat (since Clinton is
identified with his party's candidate).  The characteristic that
Common Topic relations share is that they require the identification
of parallel entities (i.e., the $a_i$ and $b_i$) and relations
($p_{0}$ and $p_{1}$) as arguments to the constraints.  We posit
that the syntactic representation is used both to guide the
identification of parallel elements and to retrieve their
semantic representations.

\subsection{Coherent Situation Inference}

Understanding utterances standing in a {\it Coherent
Situation} relation requires that hearers convince themselves that the
utterances describe a coherent situation given their knowledge of the
world.  This process requires that a path of inference be established between
the situations (i.e., events or states) described in the participating
utterances as a whole, without regard to any constraints on
parallelism between sub-sentential constituents.
Four such relations are summarized in
Table~\ref{coherent-situation-table}.\footnote{These relations are what Hume
might have termed {\it Cause or Effect}.}  In all four cases, the hearer is to
infer $A$ from sentence $S_1$ and $B$ from sentence $S_2$
under the constraint that the presuppositions
listed be abduced \cite{HobbsEtAl:93a}:\footnote{We are using
implication in a very loose sense here,
as if to mean ``could plausibly follow from''.}

\begin{table}[ht]
\begin{center}
{\small
\begin{tabular}{||c|c|c|c||} \hline
Relation & Presuppose & Conjunctions \\ \hline \hline
Result & $A \rightarrow B$ & and (as a result)  \\
 & & therefore   \\ \hline
Explanation & $B  \rightarrow A$ & because  \\ \hline
Violated Expectation & $A \rightarrow \neg B$ & but  \\ \hline
Denial of Preventer & $B \rightarrow \neg A$ & even though  \\
 & & despite  \\ \hline
\end{tabular}}
\end{center}
\caption{Coherent Situation Relations}
\label{coherent-situation-table}
\end{table}
 Examples of these relations are given in sentences (\ref{CS-examples}a-d).

\eenumsentence{ \label{CS-examples}
\item Bill is a politician, and therefore he's dishonest. (Result)
\item Bill is dishonest because he's a politician.  (Explanation)
\item Bill is a politician, but he's honest.  (Violated Expectation)
\item Bill is honest, even though he's a politician. (Denial of Preventer)
}
Beyond what is asserted by the two clauses individually, understanding
each of these sentences requires the presupposition that {\it being a
politician implies being dishonest}.  Inferring this is only reliant
on the sentential-level semantics for the clauses as a whole;
there are no $p$,
$a_i$, or $b_i$ to be independently identified. The same is true for what Hume
called {\it Contiguity relations} (perhaps including Hobbs' {\it
Occasion} and {\it Figure-ground} relations); for the purpose of this
paper we will consider these as weaker cases of {\it Cause or Effect}.

To reiterate the crucial observation, Common Topic inference utilizes
the syntactic structure in identifying the semantics for the
sub-sentential constituents to serve as arguments to the coherence
constraints.  In contrast, Coherent Situation inference utilizes only
the sentential-level semantic forms as is required for abducing a
coherent situation.  The question then arises as to what happens when
constituents in the syntax for an utterance are empty.  Given that the
discourse inference mechanisms retrieve semantic forms through nodes
in the syntax, this syntax will have
to be recovered when a node being accessed is missing.  Therefore, we
posit that missing constituents are recovered as a by-product of Common
Topic inference, to allow the parallel properties and entities serving
as arguments to the coherence relation to be accessed from within the
reconstructed structure.  On the other hand, such copying is not
triggered in Coherent Situation inference, since the arguments are
retrieved only from the top-level sentence node, which is always
present.  In the next section, we show how this difference accounts
for the data given in Section~2.

\section{Applying the Analysis}
\label{analysis-section}

In previous sections, we have classified several elliptical and event
referential forms as to whether they leave behind an empty constituent in
the syntax and whether they are anaphoric  in the
semantics.  Empty constituents in the syntax are not in themselves
referential, but are recovered during Common Topic inference.
Anaphoric expressions in the semantics are independently referential
and are resolved through purely semantic means regardless of the type
of discourse inference.  In this section we show how the
phenomena presented in Section~2 follow from these properties.

\subsection{Local Ellipsis}
\label{gapping-section}

Recall from Section~2 that
gapping constructions such as (\ref{gap1again}) are only felicitous
with the symmetric (i.e., Common Topic) meaning of {\it and}:

\enumsentence{Bill became upset, and Hillary angry. \label{gap1again}}
This fact is predicted by our account in the following way.  In the
case of Common Topic constructions, the missing sentence in the target
will be copied from the source, the sentential semantics may be
derived, and the arguments to the coherence relations can be
identified and reasoning carried out, predicting felicity.  In the
case of Coherent Situation relations, no such recovery of the syntax
takes place.  Since a gapped clause in and of itself has no sentence-level
semantics, the gapping fails to be felicitous in these cases.

This account also explains similar differences in felicity for other
coordinating conjunctions as discussed in Kehler
\shortcite{Kehler:94a}, as well as why gapping is
infelicitous in constructions with subordinating conjunctions
indicating Coherent Situation relations, as exemplified in
(\ref{othergap1}).

\enumsentence{\# Bill became upset,  $\left\{ \begin{tabular}{c}
 because \\ even though \\ despite the fact that \end{tabular} \right\}$
 Hillary angry. \label{othergap1}}

The {\it stripping} construction is similar to gapping except that there
is only one bare constituent in the target (also generally receiving
contrastive accent); unlike VP-ellipsis there is no stranded
auxiliary.  We therefore might predict that stripping is also
acceptable in Common Topic constructions but not in Coherent Situation
constructions, which appears to be the case:\footnote{Stripping is also
possible in comparative deletion constructions.  A comprehensive
analysis of stripping, pseudo-gapping, and VP-ellipsis in such cases
requires an articulation of a syntax and semantics for these
constructions, which will be carried out in future work. }

\enumsentence{Bill became upset,   $\left\{ \begin{tabular}{lc}
& and also \\ & but not \\   \# & and (as a result) \\ \# & because \\ \# &
even
though \\ \# & despite the fact that
\end{tabular} \right\}$ Hillary. \label{strip1} }

In summary, gapping and related constructions are infelicitous in those
cases where Coherent Situation inference is employed, as there is no
mechanism for recovering the sentential semantics of the elided clause.

\subsection{VP-Ellipsis}
\label{ellipsis-section}

Recall from Section~2 that only in Coherent Situation
constructions can VP-ellipsis obtain purely semantic antecedents
without regard to constraints on structural parallelism, as
exemplified by the voice mismatches in sentences (\ref{ell1again}) and
(\ref{ell2again}).

\enumsentence{\# The decision was reversed by the FBI, and the ICC did
too.  [ reverse the decision ]  \label{ell1again}}

\enumsentence{In March, four fireworks manufacturers asked that the decision be
reversed, and on Monday the ICC did. [ reverse the decision ]
\label{ell2again}}
These facts are also predicted by our account.  In the case of Common
Topic constructions, a suitable syntactic antecedent must be
reconstructed at the site of the empty VP node, with the result that
the anaphoric expression takes on its accompanying semantics.
Therefore, VP-ellipsis is predicted to require a suitable syntactic
antecedent in these scenarios.  In Coherent Situation constructions,
the empty VP node is not reconstructed.  In these cases the anaphoric
expression is resolved on purely semantic grounds; therefore VP-ellipsis is
only constrained to having a suitable semantic antecedent.

The analysis accounts for the range of data given in Kehler
\shortcite{Kehler:93b},
although one point of departure exists between that
account and the current one with respect to clauses conjoined with
{\it but}.  In the previous account these cases are all classified as
{\it non-parallel}, resulting in the prediction that they only require
semantic source representations.  In our analysis, we expect cases of
pure {\it contrast} to pattern with the {\it parallel} class since
these are Common Topic constructions; this is opposed to the {\it
violated expectation} use of {\it but} which indicates a Coherent
Situation relation.  The current account makes the
correct predictions; examples (\ref{k1}) and (\ref{k2}), where {\it
but} has the {\it contrast} meaning, appear to be markedly less
acceptable than examples (\ref{k3}) and (\ref{k4}), where {\it but}
has the {\it violated expectation} meaning:\footnote{These examples
have been adapted from several in Kehler \shortcite{Kehler:93b}.}

\enumsentence{?? Clinton was introduced by John, but Mary didn't.
[~introduce Clinton~] \label{k1}}

\enumsentence{?? This letter provoked a response from Bush, but
Clinton didn't. [~respond~] \label{k2}}
\enumsentence{Clinton was to have been introduced by someone, but
obviously nobody
did. [~introduce Clinton~] \label{k3}}

\enumsentence{This letter deserves a response, but before you do, ...
[~respond~]  \label{k4} }
To summarize thus far, the data presented in the earlier account
as well as examples that
conflict with that analysis are all predicted by the account given here.

As a final note, we consider the interaction between VP-ellipsis and
gapping.  The following pair of examples are adapted from those of Sag
\shortcite[pg. 291]{Sag:76a}:

\enumsentence{John supports Clinton, and Mary $\emptyset$ Bush, although
she doesn't know why she does. \label{sag1}}

\enumsentence{?? John supports Clinton, and Mary $\emptyset$ Bush, and
Fred does too. \label{sag2} }
Sag defines an {\it alphabetic variance} condition that correctly predicts
that sentence (\ref{sag2}) is infelicitous, but incorrectly predicts
that sentence (\ref{sag1}) is also.  Sag then suggests a weakening of his
condition, with the result that both of the above
examples are incorrectly predicted to be acceptable;  he doesn't
consider a solution predicting  the judgements  as stated.

The felicity of sentence (\ref{sag1}) and the infelicity of sentence
(\ref{sag2}) are exactly what our account predicts.  In example
(\ref{sag2}), the third clause is in a Common Topic relationship with
the second (as well as the first) and therefore requires that the VP
be reconstructed at the target site.  However, the VP is not in a
suitable form, as the object has been abstracted out of it (yielding a
trace assumption).  Therefore, the subsequent VP-ellipsis fails to be
felicitous.  In contrast, the conjunction {\it although} used before
the third clause in example (\ref{sag1}) indicates a Coherent
Situation relation.  Therefore, the VP in the third clause need not be
reconstructed, and the subsequent semantically-based resolution of the
anaphoric form succeeds.  Thus, the apparent paradox between examples
(\ref{sag1}) and (\ref{sag2}) is just what we would expect.

\subsection{Event Reference}
\label{event-reference-section}

Recall that Sag and Hankamer
\shortcite{Sag:84a} note that whereas
elliptical sentences such as (\ref{sh1}a) are unacceptable due to a
voice mismatch, similar examples with event referential forms are much
more acceptable as exemplified by sentence (\ref{sh1}b):\footnote{Sag and
Hankamer
claim that all such cases of VP-ellipsis require syntactic
antecedents, whereas we suggest that in Coherent Situation relations
VP-ellipsis operates more like their {\it Model-Interpretive
Anaphora}, of which {\it do it} is an example.}

\eenumsentence{  \label{sh1}
\item \# The decision was reversed by the FBI, and the ICC did
too.  [ reverse the decision ]
\item The decision was reversed by the FBI, and the ICC did
it too.  [ reverse the decision ] }
As stated earlier, forms such as {\it do it} are anaphoric, but leave no empty
constituents in the syntax.  Therefore, it follows under the present
account that such reference is successful without regard to the type
of discourse inference employed.

\section{Relationship to Past Work}
\label{past-work-section}

The literature on ellipsis and event reference is voluminous, and so
we will not attempt a comprehensive comparison here. Instead, we
briefly compare the current work to three previous studies that
explicitly tie ellipsis resolution to an account of discourse
structure and coherence, namely our previous account
\cite{Kehler:93b} and the accounts of Pr\"{u}st \shortcite{Prust:92}
and Asher \shortcite{Asher:93}.

In Kehler \shortcite{Kehler:93b}, we presented an analysis of
VP-ellipsis that distinguished between two types of relationship
between clauses, {\it parallel} and {\it non-parallel}.  An
architecture was presented whereby utterances were parsed into
propositional representations which were subsequently integrated into
a discourse model.  It was posited that VP-ellipsis could access
either propositional or discourse model representations: in the
case of parallel constructions, the source resided in the
propositional representation; in the case of non-parallel
constructions, the source had been integrated into the discourse
model.  In Kehler
\shortcite{Kehler:94a}, we showed how this architecture also accounted
for the facts that Levin and Prince noted about gapping.

The current work improves upon that analysis in several respects.
First, it no longer needs to be posited that syntactic representations
disappear when integrated into the discourse model;\footnote{This
claim could be dispensed with in the treatment of VP-ellipsis,
perhaps at the cost of some degree of theoretical inelegance.
However, this aspect was crucial for handling the gapping data, since
the infelicity of gapping in non-parallel constructions hinged on
there no longer being a propositional representation available as a
source.} instead, syntactic and semantic representations co-exist.
Second, various issues with regard to the interpretation of
propositional representations are now rendered moot.  Third, there is
no longer a dichotomy with respect to the level of representation from
which VP-ellipsis locates and copies antecedents.  Instead, two
distinct factors have been separated out: the resolution of missing
constituents under Common Topic inference is purely syntactic whereas
the resolution of anaphoric expressions in all cases is purely
semantic; the apparent dichotomy in VP-ellipsis data arises out of the
interaction between these different phenomena.  Finally,
the current approach more readily scales up to more complex cases.  For
instance, it was not clear in the previous account how non-parallel
constructions embedded within parallel constructions would be handled,
as in sentences (\ref{complex-ex}a-b):

\eenumsentence{ \label{complex-ex}
\item Clinton was introduced by John because Mary had refused to,
and Gore was too.  [~introduced by John because Mary had refused to~]
\item \# Clinton was introduced by John because Mary had refused to,
and Fred did too.  [ introduced Clinton because Mary had refused to~]
}
The current approach accounts for these cases.

The works of Pr\"{u}st \shortcite{Prust:92} and Asher
\shortcite{Asher:93} provide analyses of
VP-ellipsis\footnote{In addition, Pr\"{u}st  addresses gapping, and Asher
addresses event reference.} in the context of an account of discourse
structure and coherence.  With Pr\"{u}st utilizing a mixed
representation (called {\it syntactic/semantic structures}) and Asher
utilizing Discourse Representation Theory constructs, each defines
mechanisms for determining relations such as parallelism and contrast,
and gives constraints on resolving VP-ellipsis and related forms
within their more general frameworks.  However, each essentially
follows Sag in requiring that elided VP representations be alphabetic
variants of their referents.  This constraint rules out cases where
VP-ellipsis obtains syntactically mismatched antecedents, such as
example (\ref{ell2again}) and other non-parallel cases given in Kehler
\shortcite{Kehler:93b}.  It also appears that neither approach
can account for the infelicity of mixed gapping/VP-ellipsis
cases such as sentence (\ref{sag2}).

\section{Conclusion}
\label{conclusion}

In this paper, we have categorized several forms of ellipsis and event
reference according to two features: (1) whether the form leaves
behind an empty constituent in the syntax, and (2) whether the form is
anaphoric in the semantics.  We have also described two forms of
discourse inference, namely {\it Common Topic} inference and {\it
Coherent Situation} inference.  The interaction between the two
features and the two types of discourse inference predicts facts
concerning gapping, VP-ellipsis, event reference, and interclausal
coherence for which it is otherwise difficult to account.  In future
work we will address other forms of ellipsis and event reference, as
well as integrate a previous account of strict and sloppy ambiguity into
this framework \cite{Kehler:93a}.

\section*{Acknowledgments}

This work was supported in part by National Science Foundation Grant
IRI-9009018, National Science Foundation Grant IRI-9350192, and a
grant from the Xerox Corporation.  I would like to thank Stuart
Shieber, Barbara Grosz, Fernando Pereira, Mary Dalrymple, Candy
Sidner, Gregory Ward, Arild Hestvik, Shalom Lappin, Christine Nakatani,
Stanley Chen, Karen Lochbaum, and two anonymous reviewers for valuable
discussions and comments on earlier drafts.

\end{document}